# Uncertainty estimations methods for a deep learning model to aid in clinical decision-making – a clinician's perspective

Michael Dohopolski[1], Kai Wang[1], Biling Wang[1], Ti Bai[1], Dan Nguyen[1], David Sher[1], Steve Jiang[1], Jing Wang[1]

[1] Medical Artificial Intelligence and Automation Laboratory and Department of Radiation Oncology, UT Southwestern Medical Center, Dallas TX 75235, USA

michael.dohopolski@utsouthwestern.edu
github: https://github.com/MikeDoho/FT_Uncertainty_Comparison

**Abstract**

Prediction uncertainty estimation has clinical significance as it can potentially quantify prediction reliability. Clinicians may trust "blackbox" models more if robust reliability information is available, which may lead to more models being adopted into clinical practice. There are several deep learning-inspired uncertainty estimation techniques, but few are implemented on medical datasets—fewer on single institutional datasets/models. We sought to compare dropout variational inference (DO), test-time augmentation (TTA), conformal predictions, and single deterministic methods for estimating uncertainty using our model trained to predict feeding tube placement for 271 head and neck cancer patients treated with radiation. We compared the area under the curve (AUC), sensitivity, specificity, positive predictive value (PPV), and negative predictive value (NPV) trends for each method at various cutoffs that sought to stratify patients into "certain" and "uncertain" cohorts. These cutoffs were obtained by calculating the percentile "uncertainty" within the validation cohort and applied to the testing cohort. Broadly, the AUC, sensitivity, and NPV increased as the predictions were more "certain"—ie lower uncertainty estimates. However, when a majority vote (implementing 2/3 criteria: DO, TTA, conformal predictions) or a stricter approach (3/3 criteria) were used, AUC, sensitivity, and NPV improved without a notable loss in specificity or PPV. Especially for smaller, single institutional datasets, it may be important to evaluate multiple estimations techniques before incorporating a model into clinical practice.

## 1  Introduction

Novel algorithms, such as convolutional neural networks, have been able to diagnose (predict) disease states or other medically important outcomes as well as respective experts [1,2]. Despite the successes of these models, we rarely see them implemented in clinical practice. This phenomenon is multifactorial, but one concern is the validity of a particular prediction. Can clinicians trust a prediction, and how can this be measured? Can a model self-identify cases where it does not "know" if its prediction is reliable— i.e. can it say "I do not know?" Uncertainty estimation is a method proposed to quantify the reliability of a prediction and might provide a means to reassure physicians regarding a model's prediction. Alternatively, if the model can comment that it is "unsure" then the physician can ignore the prediction and rely entirely on their clinical judgment.

There are two primary sources of uncertainty: aleatoric and epistemic uncertainties [3]. At a basic level, aleatoric uncertainty is associated with inherent noise within the data; clinically, a contributor of aleatoric uncertainty may be artifact present on a CT image. Epistemic uncertainty represents a model's lack of knowledge. For example, a model trained to predict outcomes associated with a diverse array of head and neck cancers may be "uncertain" with expected to make a prediction on a poorly represented subset. Concretely, say a model was trained on 100 oropharyngeal cancer cases, 80 laryngeal cancer cases, and three nasopharyngeal cancer cases. When asked to make another prediction on a new nasopharyngeal cancer case, it might be more "uncertain" compared to a new oropharyngeal cancer case. Alternatively, epistemic might be able to identify model misuse, where a prediction made by the model above would be more "uncertain" if exposed to a thyroid cancer case. While limited, uncertainty estimation has shown an increased prevalence in the medical-based machine learning literature, and there have been several proposed methods for estimating both aleatoric and epistemic uncertainty [3–9].

Most comparison studies on uncertainty estimation methods describe the pros and cons associated with various methods [10]. For example, dropout variation inference and test-time augmentation methods are more computationally expensive when making predictions on test data versus ensemble methods, which alternatively require significant computational effort during training. Single deterministic approaches are less computationally expensive than the prior methods but require the model to be retrained to predict the uncertainty distribution [5,6,8,9,11].

Excitingly, Berger et al. recently published their work, which compared out-of-distribution detection methods (i.e. epistemic uncertainty surrogates) using a large publicly available medical dataset (CheXpert) and made the critical observation that associated performance on traditional computer vision datasets do not always translate as well when applied to medical dataset [4,12]. They also briefly explored threshold selection for out-of-distribution identification using temperature scaling—high confidence/low uncertainty was associated with higher accuracy. However, accuracy is only one metric that is important to a clinician; sensitivity and specificity are key. Moreover, threshold selection is critical as having too strict a threshold might limit a model's utility (i.e. decrease the patient sample size where the model is "certain") or negatively affect clinically used metrics such as sensitivity or specificity [13].

This study employs several epistemic and aleatoric uncertainty estimation methods and compares performance at various cutoffs using AUC, sensitivity, and specificity for a model trained to predict a clinically significant event—feeding tube placement— in head and neck cancer patients treated with definitive radiation therapy. Feeding tube placement is important for nutritional supplementation. If patients are not accurately identified that they may need a feeding tube, then treatment delays can occur, which are associated with worse survival [14,15]. Meanwhile, feeding tube placement is a surgical procedure and can be associated with worse quality of life. So accurately predicting which patients need a feeding tube is important [16]. Our particular clinical scenario aside, many medical dilemmas require reliable criteria to make proper medical decision-making. We hope our uncertainty estimation analyses highlight practical challenges in implementing these methods for models trained on relatively small single-institutional datasets. In conducting our analyses, we introduce the idea of implementing multiple uncertainty estimation methods to improve general discriminative ability while not sacrificing other metrics like specificity or sensitivity—an application we have not previously seen.

## 2    Methods

### 1.    Data

This single institutional dataset included 271 patients. The outcome predicted was feeding tube placement or ≥10% weight loss if the patient declined feeding tube placement; this accounted for 42% of patients within the dataset. CT imaging, including the radiation planning CT and on-treatment cone-beam CT, and radiation dose were used as a three-channel input. The input was of size $150 \times 80 \times 80$ to focus on the oral cavity, oropharynx, and esophagus (structures important for swallowing) [17].

### 2.    Model

Transfer learning using MedNet's ResNet50's architecture was employed [18,19]. Fivefold cross-validation was performed. Roughly 80% and 20% were used for training/validation and testing. The original model was trained with stochastic gradient descent using a momentum of 0.9 and weight decay of 0.001. Batch normalization and dropout were utilized. The model was trained over 120 epochs without early stopping. Cross entropy with a class weight of 1/3 and 2/3 was used for class 0 and class 1. A separate model was trained using the same hyperparameters but a different loss function that incorporated a Kullback-Leibler divergence term to create the deterministic/evidential deep learning model [20].

### 3.    Uncertainty Estimation Methods

**Dropout Variational Inference**
The implementation of dropout (DO) variational inference was popularized by Kendall and Gal to approximate epistemic uncertainty. A model is trained with dropout and incorporates dropout at test time. Multiple predictions are made for a single image, and the class probabilities are used to calculate the informational entropy [9]. We used 300 repeated predictions with a dropout rate of 0.05. A dropout layer was added to each convolutional layer [19].

**Test-time Augmentation**

Test-time augmentation (TTA) seeks to estimate aleatoric uncertainty. This technique makes repeated predictions on the same image after different data augmentations are applied during test time. The predictive distribution is obtained, and the informational entropy is calculated [5]. We used 300 repeated predictions. One to three augmentations (shift, rotation, blur) were randomly selected.

**Conformal Predictions**

Conformal prediction assigns an input into a set of possible labels. Instead of a class probability, a prediction could include no label, a single label, or up to the maximal number of labels. The algorithm's goal is to classify test data into various classes using a calibration set. We used the nonconformist python library to create an inductive conformal classifier [21,22]. Next, we calculated a prediction's "credibility," which is the minimal significance level that the conformal region is empty, to calculate how likely an input came from a similar distribution as the training set.

**Evidential Deep Learning**

Evidential deep learning (EvDL) is a deterministic approach for measuring epistemic uncertainty. In EvDL, for the classification tasks, Dirichlet priors are placed over the likelihood, and the model is trained to learn the alpha parameters. The model is trained with a loss function incorporating a Kullback-Leibler divergence term. The uncertainty is calculated using the number of class labels divided by the Dirichlet strength [11,20].

## 4. Metrics and analyses

AUC, sensitivity, specificity, positive predictive value (PPV), and negative predictive value (NPV) were calculated for each cutoff value. The percent change was calculated based on the baseline performance metric for the whole, unstratified cohort. Unique cutoff values were calculated using various percentile values from the validation dataset for each uncertainty estimation method and used to stratify prediction into "certain" and "uncertain" cohorts. For example, "uncertain" predictions were those greater than or equal to a particular percentile cutoff, which could range from the 30th to 90$^{th}$ percentile. The validation values were used on the testing data. Separately, "credibility" cutoff values ranging from 0.30 to 0.90 were used for conformal predictions—no percentile calculations were performed. The percent of patients in a particular cohort was calculated using these cutoff values. We also investigated employing all TTA, DO, and conformal prediction criteria (3/3 criteria, "ALL") and a majority voting strategy (if met 2/3 criteria, "majority"). We also evaluated where each uncertainty estimation technique increased the validation AUC by 10%. 95% confidence intervals were calculated at each cutoff.

## 3   Results

The baseline AUC for the model that used TTA, DO, and conformal prediction as uncertainty estimates was 0.72 and 0.72 for validation and testing, respectively. The baseline AUC for the model that used EvDL was 0.69 and 0.67 for validation and testing, respectively. As predictions became more "certain"—ie less percentage of patients within the low uncertainty cohort—the AUC within the validation cohort improved compared to the overall/baseline performance (Figure 1A). A 10% improvement in the validation AUC was seen when the "certain" cohort retained 40% of the total patient population. Majority, ALL, and TTA retained ~30-35% of the total population before noting a 10% increase in AUC. In the testing data (Figure 1B), a rising AUC trend was observed sooner (~80%). For predictions with increasing "uncertainty," it was difficult to appreciate a trend in AUC for the various methods aside from DO and EvDL, which showed a slight downtrend in the validation data (Figure 1C). In the testing dataset (Figure 1D), all methods noted a more notable downward trend in AUC as the cohort was more "uncertain."

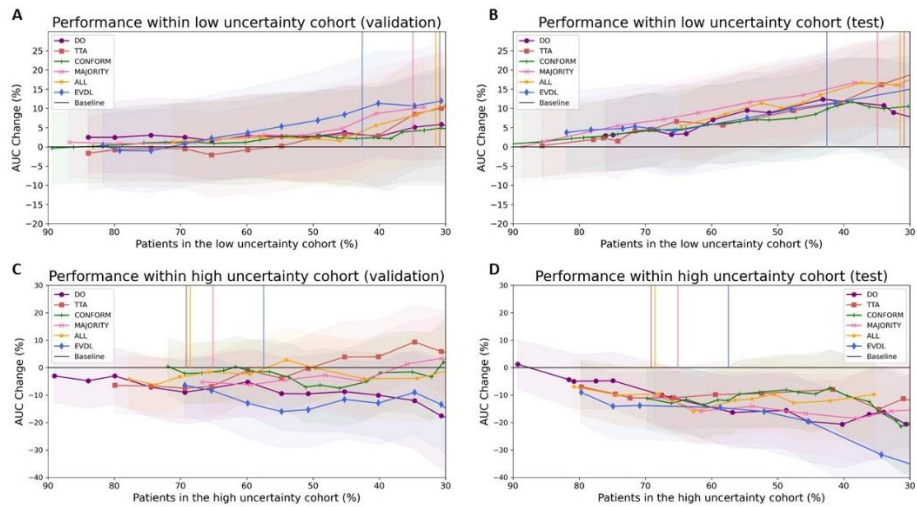

**Fig. 1.** A) The AUC change as the cohort became more "certain," ie, the percentage of patients within the low uncertainty group decreased within the validation dataset. B) The AUC change as the cohort became more "certain" within the testing dataset. C) The AUC change as the cohort became more "uncertain," ie, the percentage of patients within the high uncertainty group decreased within the validation dataset. D) The AUC change as the cohort became more "uncertain" within the test dataset. Abbreviations: All – applied DO, TTA, conformal prediction criteria (3/3 criteria); Majority – applied 2/3 DO, TTA, conformal prediction criteria.

The sensitivity improved as predictions within the validation dataset were more "certain" (Figure 2A). Notably, the EvDL and TTA methods showed lower than baseline performance. However, most methods showed notable improvement within the testing dataset as predictions became more "certain." The early trends for EvDL and TTA were not as promising as the other methods but illustrated a positive trend when looking at the cutoff associated with a 10% improvement in AUC (Figure 2B). Interestingly, the specificity within the low uncertainty validation cohort had inverse performance trends compared to sensitivity for DO, EvDL, TTA, and conformal prediction. A majority voting or stricter approach seemed to blunt any notable decline in specificity as sensitivity improved (Figure 2C/D).

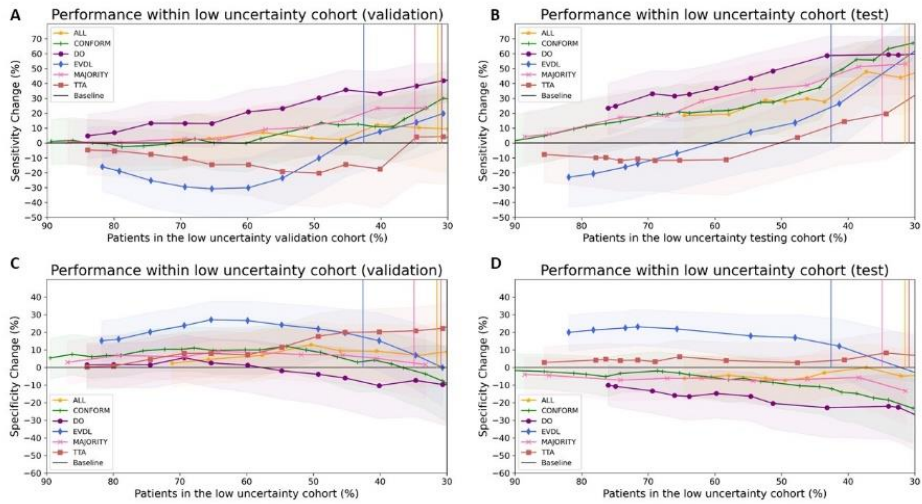

**Fig. 2.** A) The sensitivity change as the cohort became more "certain," ie, the percentage of patients within the low uncertainty group decreased within the validation dataset. B) The sensitivity change as the cohort became more "certain" within the testing dataset. C) The specificity change as the cohort became more "uncertain," ie, the percentage of patients within the high uncertainty group decreased within the validation dataset. D) The specificity change as the cohort became more "uncertain" within the test dataset. Abbreviations: All – applied DO, TTA, conformal prediction criteria (3/3 criteria); Majority – applied 2/3 DO, TTA, conformal prediction criteria.

The sensitivity remained unchanged or decreased as predictions were more "uncertain" for conformal prediction, ALL, DO, and a majority (Figure 3A). The EvDL and TTA methods showed improved sensitivity. Most methods aside from TTA and EvDL showed worse sensitivity within "uncertain" test cohorts (Figure 3B). The specificity within the uncertain validation cohorts decreased as the uncertainty estimates increased for the majority, conformal prediction, and EvDL. ALL and DO methods showed minimal change in specificity (Figure 3C). All methods aside from TTA and EvDL showed improved specificity within the "uncertain" testing cohorts (Figure 3D).

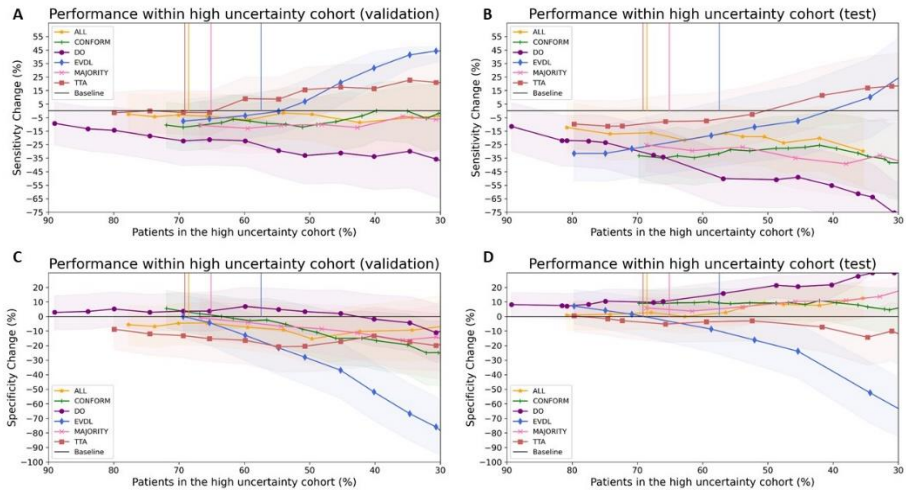

**Fig. 3.** A) The sensitivity change as the cohort became more "uncertain," ie, the percentage of patients within the high uncertainty group decreased within the validation dataset. B) The sensitivity change as the cohort became more "uncertain" within the testing dataset. C) The specificity change as the cohort became more "uncertain," ie, percentage of patients within the high uncertainty group decreased within the validation dataset. D) The specificity change as the cohort became more "uncertain" within the test dataset. Abbreviations: All – applied DO, TTA, conformal prediction criteria (3/3 criteria); Majority – applied 2/3 DO, TTA, conformal prediction criteria.

The PPV in the validation and test datasets did not appreciably change as predictions became more "certain" (Figure 4A/B) Interestingly, EvDL had an overall unchanging increase in the testing dataset. NPV increased in the validation and testing dataset for all methods (Figure 4C/D).

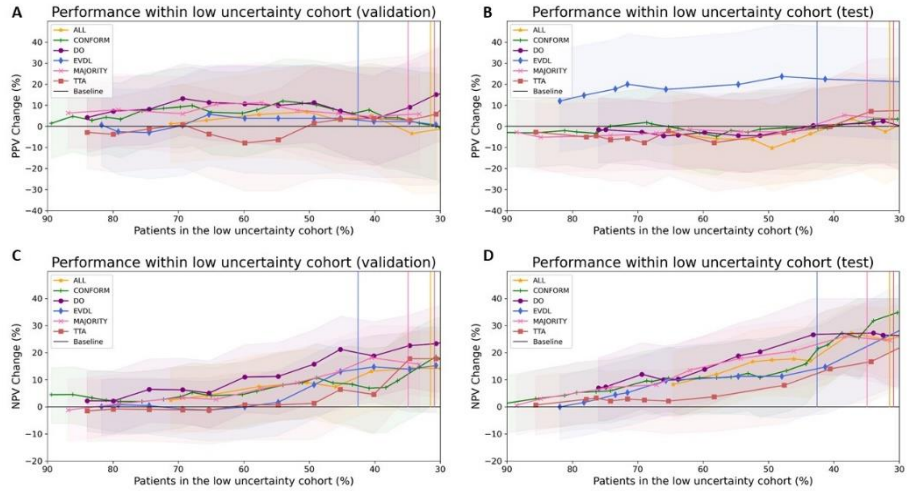

**Fig. 4.** A) The PPV change as the cohort became more "certain," ie, the percentage of patients within the low uncertainty group decreased within the validation dataset. B) The PPV change as the cohort became more "certain" within the testing dataset. C) The specificity change as the cohort became more "certain," ie, the percentage of patients within the low uncertainty group decreased within the validation dataset. D) The specificity change as the cohort became more "certain" within the test dataset. Abbreviations: All – applied DO, TTA, conformal prediction criteria (3/3 criteria); Majority – applied 2/3 DO, TTA, conformal prediction criteria.

The PPV in the validation and test datasets did not appreciably change as predictions became more "uncertain" (Figure 5), aside from DO, which increased in the testing dataset. NPV remained unchanged in the validation dataset for all methods (Figure 5) and remained relatively unchanged and lower than the baseline performance in the test dataset aside from EvDL, which had a downward trend.

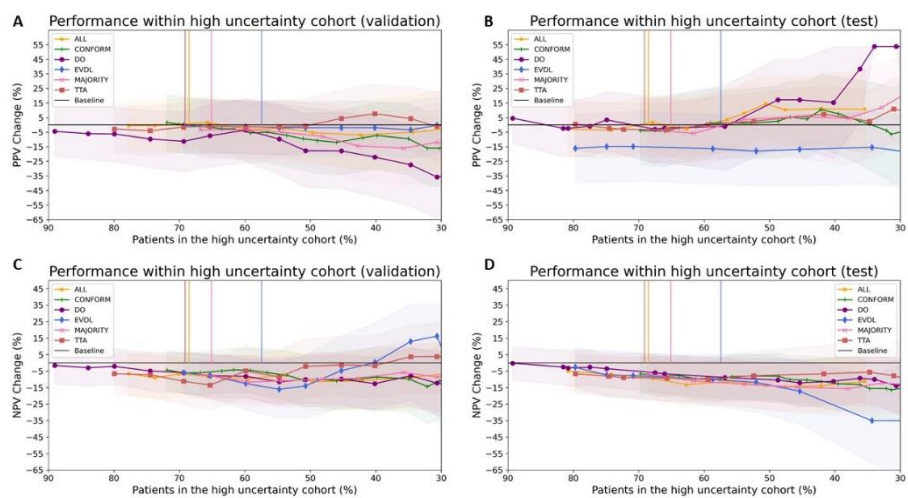

**Fig. 5.** A) The PPV change as the cohort became more "uncertain," ie, the percentage of patients within the high uncertainty group decreased within the validation dataset. B) The PPV change as the cohort became more "uncertain" within the testing dataset. C) The NPV change as the cohort became more "uncertain," ie, the percentage of patients within the high uncertainty group decreased within the validation dataset. D) The NPV change as the cohort became more "uncertain" within the test dataset. Abbreviations: All – applied DO, TTA, conformal prediction criteria (3/3 criteria); Majority – applied 2/3 DO, TTA, conformal prediction criteria.

The outcome percentage (with and without feeding tube placement) was similar to the baseline cohort (42%) for all validation and testing "certain" and "uncertain" cohorts for each cutoff. The percentages (mean/std) for patients with feeding tube placement in the low uncertainty testing (mean/std) – DO: 45%/3%, TTA: 37%/2%, conformal prediction: 43%/4%, EvDL: 43%/5%, majority: 42%/2%, ALL: 37%/2%. Similar trends were in the validation low and all high uncertainty cohorts.

## 4    Conclusion

Accurate predictions are critical in medicine as they guide treatment decisions, which can have lasting complications [23]. Our model was trained to recognize patients who were going to need an invasive surgical procedure (feeding tube placement) to aid in nutritional replacement. Patients who will truly need a feeding tube can have this procedure scheduled and not have their current radiation therapy delayed. Delays in radiation therapy for head and neck cancers increase mortality [14,15]. However, for those who will tolerate treatment without a feeding tube, unnecessary placement exposes the patient to potential complications and quality of life deficits. This is one example of why accurate predictions in medicine are so critical. Yet, a physician must have confidence in a model's prediction to trust an algorithm for clinical decision-making. Uncertainty quantification is an attractive method that can potentially identify patients where the model might make more accurate predictions [24]. Physicians might be more willing to trust a blackbox model if sufficient data supports that a model makes reliable, single predictions within a particular cohort of patients [6].

Here we explored several methods that estimate uncertainty to observe any performance benefit or improved reliability in a progressively more "certain" cohort. Evaluating only AUC, several methods showed that more "certain" predictions were generally more accurate; however, increased certainty came at the expense of the model making predictions on a much smaller cohort of patients. The fewer patients the model will report a prediction (the more times the model says "I do not know"), the less useful the model is in the clinic. We set our cutoff for each method as the value associated with a 10% improvement in AUC in the validation dataset. EvDL, TTA, majority, ALL met these criteria. EvDL did so while retaining the most patients in the "certain" cohort—~40%.

AUC may be difficult to use directly for clinical decision-making. Physicians prioritize model sensitivity, specificity, PPV, and NPV, given vastly different implications in treatment pathways. For example, suppose a model is predicting cancer recurrence. In that case, we do not want to miss any new or residual tumors as they become more challenging to treat, so sensitivity, in this case, is important. In our example, it was exciting to observe that sensitivity and NPV improved in the more "certain" cohorts. However, for many methods, this was at the expense of specificity. None of these methods displayed a trend for increased general performance (AUC) while also increasing specificity or sensitivity without significantly decreasing the other metric. However, when a majority vote or stricter approach (ALL) was used, trends that appeared more reliable were noted in the low uncertainty validation and testing cohorts (i.e. AUC and sensitivity improved without significant loss in specificity). Therefore, it is likely important to evaluate several methods of uncertainty estimation when potentially incorporating a model into clinical practice.

There are several limitations in this work. One limitation is that we had two baseline models—one used for DO, TTA, and conformal predictions and one for EvDL. We could have applied the above methods to the EvDL model, but its baseline performance was worse than the original model; thus, we kept the analyses separate. The most significant limitation, however, was that these analyses were done using only a few estimation methods (specifically missing a popular method—model ensembling) and on a relatively small, single institutional dataset. However, many medical datasets are small. These general analyses might prompt others to make comprehensive uncertainty estimation analyses for the models they might use at their institutions. In the future, we hope to explore the clinical and imaging characteristics of these "uncertain" patients to see how these patients differ from the "certain" cohort.